# Ground Material Segmentation for UAV-based Photogrammetric 3D Data: A 2D-3D Hybrid Approach

**Meida Chen, Andrew Feng, Yu Hou, Kyle McCullough, Pratusha Bhuvana Prasad,**
***U.S.C. Institute for Creative Technologies***
**Los Angeles, California**
**{mechen, feng, yuhou, McCullough, bprasad}@ict.usc.edu**

**Lucio Soibelman**
***USC Department of Civil and Environmental Engineering***
**Los Angeles, California**
**soibelman@usc.edu**

## ABSTRACT

In recent years, photogrammetry has been widely used in many areas to create photorealistic 3D virtual data representing the physical environment. The innovation of small unmanned aerial vehicles (sUAVs) has provided additional high-resolution imaging capabilities with low cost for mapping a relatively large area of interest. These cutting-edge technologies have caught the US Army and Navy's attention for the purpose of rapid 3D battlefield reconstruction, virtual training, and simulations. Our previous works have demonstrated the importance of information extraction from the derived photogrammetric data to create semantic-rich virtual environments (Chen et al., 2019). For example, an increase of simulation realism and fidelity was achieved by segmenting and replacing photogrammetric trees with game-ready tree models. In this work, we further investigated the semantic information extraction problem and focused on the ground material segmentation and object detection tasks. The main innovation of this work was that we leveraged both the original 2D images and the derived 3D photogrammetric data to overcome the challenges faced when using each individual data source. For ground material segmentation, we utilized an existing convolutional neural network architecture (i.e., 3DMV) which was originally designed for segmenting RGB-D sensed indoor data. We improved its performance for outdoor photogrammetric data by introducing a depth pooling layer in the architecture to take into consideration the distance between the source images and the reconstructed terrain model. To test the performance of our improved 3DMV, a ground truth ground material database was created using data from the One World Terrain (OWT) data repository. Finally, a workflow for importing the segmented ground materials into a virtual simulation scene was introduced, and visual results are reported in this paper.

## ABOUT THE AUTHORS

**Meida Chen** is currently a research associate at the University of Southern California's Institute for Creative Technologies (USC-ICT) working on One World Terrain project. He received his Ph.D. degree at USC. Sonny Astani Department of Civil and Environmental Engineering. His research focuses on the semantic modeling of outdoor scenes for the creation of virtual environments and simulations. Email: mechen@ict.usc.edu

**Andrew Feng** is currently a research scientist at USC-ICT working on the One World Terrain project. Previously, he was a research associate focusing on character animation and automatic 3D avatar generation. His research work involves applying machine learning techniques to solve computer graphics problems such as animation synthesis, mesh skinning, and mesh deformation. He received his Ph.D. and MS degree in computer science from the University of Illinois at Urbana-Champaign. Email: feng@ict.usc.edu

**Yu Hou** is a Ph.D. candidate working with Dr. Lucio Soibelman. He received his M.S. degree in construction management from China University of Mining and Technology. He joined USC in 2016 and earned an M.S. degree in computer science from USC in 2019. He is expected to graduate in August 2021. His research focuses on improving building energy efficiency by automatically detecting heat loss and moisture areas on the building envelopes from 3D thermographic models reconstructed by drone-based thermal images.

**Kyle McCullough** is currently the lead Programmer at USC-ICT working on the One World Terrain project. Previously, he was a creative director and writer for the Video Games Industry, most recently for Ubisoft's 'Transference', winning Best Interactive Narrative VR Experience at Raindance 2018. His research work involves advanced prototype systems development, utilizing AI and 3D visualization to increase fidelity and realism in large-scale dynamic simulation environments. He has a BFA from New York University.
Email: McCullough@ict.usc.edu

**Pratusha Bhuvana Prasad** is currently a researcher at USC-ICT working on the One World Terrain project. Her research focuses on computer vision for geometry and using machine learning methods to solve the same. She has a Master's degree from the Ming Hsieh Department of Electrical and Computer Engineering, USC. Email: bprasad@ict.usc.edu

**Lucio Soibelman** is a Professor and Chair of the Sonny Astani Department of Civil and Environmental Engineering at USC Dr. Soibelman's research focuses on use of information technology for economic development, information technology support for construction management, process integration during the development of large-scale engineering systems, information logistics, artificial intelligence, data mining, knowledge discovery, image reasoning, text mining, machine learning, multi-reasoning mechanisms, sensors, sensor networks, and advanced infrastructure systems. Email: soibelman@usc.edu

# Ground Material Segmentation for UAV-based Photogrammetric 3D Data: A 2D-3D Hybrid Approach

**Meida Chen, Andrew Feng, Yu Hou, Kyle McCullough, Pratusha Bhuvana Prasad,**
***U.S.C. Institute for Creative Technologies***
**Los Angeles, California**
**{mechen, feng, yuhou, McCullough, bprasad}@ict.usc.edu**

**Lucio Soibelman**
***USC Department of Civil and Environmental Engineering***
**Los Angeles, California**
**soibelman@usc.edu**

## INTRODUCTION AND BACKGROUND

One of the main challenges faced by the US Army modeling & simulation (M&S) community has been the 3D modeling process, which has traditionally relied almost exclusively on the brute force approach for creating their virtual surroundings – hiring teams of artists, content producers, and programmers to create the terrain, buildings, roads, vegetation, etc. in which a simulation takes place. This virtual content production process takes tremendous resources (e.g., time, money, and manpower), in many instances cannot be reused, and often \times does not reflect reality. In recent years, photogrammetry has been widely used in many areas to create photorealistic 3D virtual data representing the physical environment. The innovation of small unmanned aerial vehicles (sUAVs) has provided additional high-resolution imaging capabilities with low cost for mapping a relatively large area of interest. These cutting-edge technologies have caught the US Army and Navy's attention for the purpose of rapid 3D battlefield reconstruction, virtual training, and simulations. However, this high-resolution 3D environment reconstruction capability has only partially addressed the challenges mentioned above.

The photogrammetry technique can be used to generate 3D meshes that represent the entire area of interest. However, it does not generate the 3D meshes with the semantic information which exists in the 3D models that were hand-created by artists. Our previous works have demonstrated the importance of information extraction from the derived photogrammetric data to create semantic-rich virtual simulation environments (Chen et al., 2019). The research prototype – Semantic Terrain Points Labeling System (STPLS) that we have developed was used to segment the photogrammetric data into top-level terrain elements (i.e., ground, man-made structures, and vegetation) and extract object information such as individual tree locations and building footprints. An increase in simulation realism and fidelity was achieved by segmenting and replacing photogrammetric trees with game-ready tree models. In this work, we further investigated the semantic information extraction problem and focused on the ground material segmentation task.

Recognizing material categories from 2D images is a fundamental problem in computer vision and has drawn attention from both academia and industry in the past two decades (Adelson, 2001). The problem is usually viewed as a texture classification problem and has been studied side by side with image classification and object detection. Earlier works have focused on investigating and developing statistical approaches to quantify handcrafted texture features that were extracted using different texture filters (e.g., Sobel filter, color histograms, filter banks, Gabor filters, etc.) for the material classification problems. Some of the proposed material classification approaches have benchmarked using a texture database that was created in a controlled environment [- i.e., CUReT dataset (Dana, Van Ginneken, Nayar, & Koenderink, 1999)] and could achieve classification accuracy over 95% (Liu, L. & Fieguth, 2012; Varma & Zisserman, 2009). However, M. Varma and A. Zisserman's approach (Varma & Zisserman, 2009) could only achieve a classification accuracy of 23.8% when applied to a database that contains common materials with real-world appearances (- i.e., Flickr Material Database – FMD) (Sharan, Rosenholtz, & Adelson, 2009). C. Liu et al. improved the classification accuracy on FMD to 44.6% in 2010 with a proposed Bayesian learning framework and a set of new features (Liu, C., Sharan, Adelson, & Rosenholtz, 2010). D. Hu et al. improved the classification accuracy again on FMD to 54% in 2011 with a set of proposed variances of gradient orientation and magnitude texture features. With the recent advancement in deep learning, researchers started to investigate and develop different Convolutional Neural Network (CNN) architectures for solving the material classification tasks. G. Kalliatakis et al. compared three CNN architectures—AlexNet, OverFeat, and a CNN architecture introduced in (Zeiler & Fergus, 2014)—for material classification problems and benchmarked using FMD. The results indicated that material classification using CNN models for FMD could achieve over 60% accuracy. M. Cimpoi et al. extracted texture features using the VGG model and performed the classification using SVM, which achieved 82.4% accuracy for FMD (Cimpoi, Maji, & Vedaldi, 2015).

Previous studies have focused on and made valuable contributions to the material classification problem for real-world 2D images. In 2019, we investigated the CNN solutions for the ground material segmentation problem with mesh rendered 2D images (Chen et al., 2019). We also showed the importance of providing such extracted information to the simulation engine for maneuverability analysis. This study further researched the problem and tried to provide a better solution by leveraging both the source images and the derived 3D photogrammetric data. On the one hand, using only the 2D images for the ground material segmentation task ignores the entire 3D geometric features when the 3D geometric features could potentially improve the segmentation result. For instance, the 3D data of a grass area could be flatter than the bare earth with rocks. On the other hand, using only the 3D data for the segmentation task ignores important color and 2D texture information that can be used to distinguish between natural surfaces of the ground (i.e., grass area) and man-made terrain surface (i.e., paved road). Thus, this study aims to utilize both 2D and 3D data to overcome the challenges faced when using each individual data source. We utilized the existing 3DMV network architecture and designed a depth pooling layer to improve its performance on the outdoor photogrammetric data. A large ground material database was created in this study and used for validation purposes. We also developed a workflow to import the extracted ground materials into the virtual environments to improve its realism and fidelity.

## GROUND MATERIAL SEGMENTATION

To better understand the ground material segmentation problem, we first investigated the strength and limitations of the existing 3D data collection techniques (e.g., terrestrial laser scanning, airborne laser scanning, and photogrammetry) for mapping large areas of interest. Photogrammetric data has the characteristics that (1) high-resolution source images that are used to reconstruct the 3D data do exist, and (2) The derived 3D point clouds are not as accurate as of the laser-scanned point clouds, e.g., small geometric details (< 10 cm) are not being captured. Considering the nature of ground photogrammetric data and the task of segmenting ground materials, an algorithm that can take advantage of the high-resolution source images and tolerate the low accuracy of the 3D data is needed.

### Model selection

There is a long history of investigation of the 3D data (i.e., point cloud) segmentation and classification. Most of the previous work has been focused on LIDAR surveyed outdoor data and RGB-D sensed indoor data. Researchers have established several benchmark databases such as Semantic3D, S3DIS, and ScanNet for comparing and evaluating different algorithms with respect to a certain performance measure (i.e., mean intersection over union – MIOU, mean accuracy – MACC, and overall accuracy – OACC). Most of the objects for classification and segmentation in the existing benchmark datasets are the ones that could be potentially better identified by their geometric shapes than their colors and textures. For instance, the Semantic3D dataset contains eight object categories, including terrain, vegetation, hardscape, scanning artifacts, cars, etc.; the S3DIS dataset contains 12 object categories including ceiling, floor, wall, beam, column, window, door, table, chair, sofa, bookcase and board; and ScanNet dataset contains 19 object categories including floor, wall, cabinet, bed, chair, sofa, table, door, window, bookshelf, picture, counter, desk, curtain, refrigerator, bathtub, shower curtain, toilet, and sink. Out of the three benchmark databases mentioned above, only a few objects can be better identified by their colors and textures such as picture, curtain, and window.

Consequently, many of the existing algorithms were designed to utilize the geometric features and use color as additional information if it exists. When applying such an algorithm (e.g., PointNet++) on the ScanNet, it was able to achieve an OACC of 60.2%. However, when we look closely at its performance on objects that do not have notable geometric features such as picture, curtain, and window, we found it was able to achieve an OACC of only 23.7% for the window, 48.7% for the curtain, and 0% for the picture. The algorithm that utilizes both the 3D geometric feature and 2D texture information such as 3DMV, on the other hand, was designed to improve the segmentation performance on these objects. 3DMV was able to achieve an OACC of 75% on ScanNet. In particular, it achieved an OACC of 61.2% for the window, 82.4% for the curtain, and 55.8%% for the picture. Thus, 3DMV was selected in this study to solve the problem of segmenting ground materials. However, since the original 3DMV algorithm was designed and validated with indoor data, it cannot be applied directly to the outdoor photogrammetry data. In the following sections, we will discuss the original 3DMV architecture along with our modifications and design decisions behind it.

### Data representation

To better understand the 3DMV model architecture, different data representations for point cloud segmentation need to be introduced first. Designing an efficient and effective 3D representation for deep learning algorithms has been and remains an active field of research. There are three main ways of representing 3D point clouds that are then fed into deep-learning segmentation algorithms: (1) volumetric representation, (2) an unordered point set (i.e., the original form of point clouds), and (3) 2D representation of the 3D data (e.g., depth map, texture map, and corresponding images or rendered images). Since 3DMV uses both (1) and (3), we will introduce these two representations in detail.

3D volumetric representation was adopted directly from the 2D image representation for segmentation and classification, where neural networks expect a fixed size 2D matrix as the input. In order to prepare the 3D point cloud in the volumetric representation, a 3D scene is divided into equally sized voxels. The 3D object geometry information is stored as occupancy value, in which a voxel has points inside is 1 and 0 otherwise. There are several other ways of computing the value of each voxel, such as computing the point density for each voxel or computing the average RGB value for each voxel. Most of the previous studies that used volumetric representation for point cloud segmentation only used the occupancy value. For dealing with a large dataset that has a high point density, fitting the entire voxel grid into a neural network may not be feasible. Thus, a common practice is to split the whole voxel grid into small chunks and perform the segmentation process on the individual chunk. In this study, we first divided the 3D point cloud into chunks where width ($W_L$), depth ($D_L$), and height ($H_L$) are the size of a chunk along each X, Y, and Z-axis. Following that, each chunk was subdivided into width ($W_S$) * depth ($D_S$) * height ($H_S$) small grid of voxels. To make computation more efficient, the chunk that does not contain any points were eliminated during the first process. For the sake of convenience, $W_L$, $D_L$, and $H_L$ were defined to be divisible by $W_S$, $D_S$, and $H_S$, respectively. The occupancy value (i.e., 0 or 1) was used in this study to represent the 3D geometry in the voxel grids.

Approaches that perform 3D point cloud segmentation using 2D images are commonly referred to as multi-view fusion. An essential part of such an approach is establishing the correspondence between the 2D images and points in the 3D space. Depending on the equipment that is being used for data collection, such correspondence may or may not exist. For instance, with LIDAR data such as the Semantic3D dataset, RGB images were not collected, and the point clouds were not colorized. Researchers tried different ways to project the 3D points onto the 2D space (e.g., Cartesian bird's-eye projection and polar bird's-eye projection) and used the 2D CNN architectures to perform the segmentation task. On the other hand, in RGB-D sensed data such as S3DIS and ScanNet, the correspondence between the collected images and the 3D point clouds already exists. Thus, the segmented 2D photos can be directly back-projected to the 3D point cloud. The photogrammetric data is similar to the RGB-D sensed data, where the correspondence also exists. One of the photogrammetry sub-processes is aerial triangulation, which is used to compute such correspondence and estimate the intrinsic and extrinsic camera parameters for each source image. ContextCapture, the photogrammetry software, was used for reconstructing the 3D point clouds in this study, and both the estimated intrinsic and extrinsic parameters were exported directly from the software.

### 3DMV Architecture

Dai, A. and Nießner, M. designed 3DMV in 2018 for segmenting RGB-D sensed indoor data and achieved state-of-the-art performance on the ScanNet benchmark data set at the year it was published (it outperformed the previous state-of-the-art algorithm by 22.2% accuracy) (Dai & Nießner, 2018). 3DMV is a joint 3D multi-view segmentation network architecture that can be trained end-to-end. The main contribution of the 3DMV work is the proposed 2D-3D differentiable back-projection layer. The 3DMV architecture is composed of a 2D and a 3D network where the 2D network extracts the features from the 2D images and then back-projects them onto the 3D voxel grids to feed into the 3D network. For more information on the original 3DMV implementation, readers can refer to (Dai & Nießner, 2018). This paper will briefly discuss each part of the network and emphasize our modifications to the network for improving its performance on outdoor photogrammetric data.

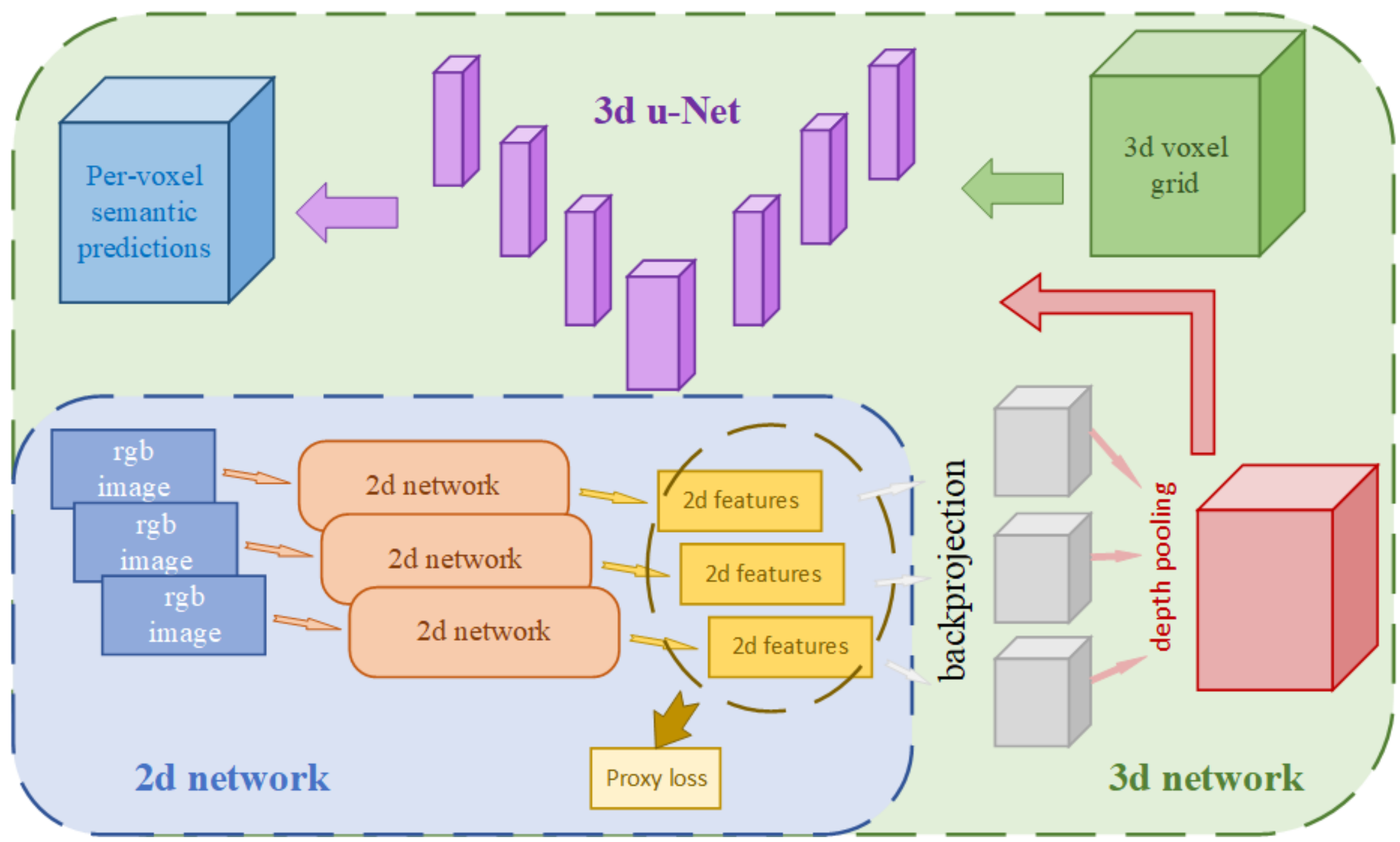

**Figure 1. Modified 3DMV network architecture. Please refer to Figure 2 in** (Dai & Nießner, 2018) **for the original 3DMV architecture.**

Our modified 3DMV architecture is shown in Figure 1. For the 3D network, u-Net architecture is used in this study. The input of the 3D U-Net is the 3D voxels with occupancy values and the back-projected 2D features that are extracted from the 2D network. The 3D voxels are first fed through several encoding units that consist of convolution layers and max-pooling layers to extract the 3D feature maps. The dimension of the intermediate feature maps is 32, 64, 128, and 256. The voxels are down-sampled through the encoding units. The lowest resolution feature map is extracted from the last encoding unit, where the dimension of the feature map is 256. Low-resolution feature maps are then recursively up-sampled to full resolution through a set of decoding units. To allow final segmentation to take place with an awareness of the high-level features in the feature maps that are extracted from the encoding units, short- and long-skip connections are also used in the network between the encoding and decoding units. Finally, the label is assigned to each voxel within the voxel grids through a 1 * 1 * 1 convolution layer. To improve training stability and efficiency, we also added a batch normalization layer between each convolution layer. Since using the model to make predictions on new unseen data is the ultimate goal, a dropout layer with a 50% dropout rate is applied after each max-pooling layer to avoid overfitting. The output layer then produces the semantic segmentation results for the given inputs (i.e., 3D voxels with label predictions). The labeled voxels will then be converted back to the point cloud format with its original point spacing, and if one voxel contains several points, all these points will be assigned to the same label that is predicted.

We have adopted the same 2D network—the ENet architecture (Paszke, Chaurasia, Kim, & Culurciello, 2016)—as the original 3DMV. Since 3DMV contains both 2D and 3D networks, training such a large network would be time-consuming and requires a large amount of GPU memory. Thus, ENet has been selected mainly because of its simplicity, as it is both fast to run and memory-efficient to train. The main goal of using the 2D network is to extract the 2D features from the source images but not segment the source images. However, to extract meaningful features and make the training process more stable, the 2D network was also trained to perform the segmentation task. A 2D proxy loss was used during the training process for the 2D network. The 2D features are extracted from the last layer before computing the segmentation scores; these feature maps are then back-projected onto the 3D voxels using a differentiable back-projection layer. In this study, the source images were down-sampled through the network to 41 by 32 pixels. Following that, 128 features were obtained for each pixel and back-projected to their corresponding 3D voxels.

We followed the original back-projection layer implementation and used the camera intrinsic and extrinsic parameters to establish the correspondence between each learned 2D feature map and the 3D voxel grids. Since this work focused on segmenting ground materials, non-ground objects, such as buildings and trees, were irrelevant. We used our previously designed Semantic Terrain Points Labeling System (STPLS+) (Chen et al., 2019) to segment the 3D data into the ground and non-ground layers and only fed the ground layer data to the 3DMV network to reduce the computational time. However, since the 2D features were extracted from the original source images, features for non-ground objects were also extracted and back-projected to the ground layer 3D voxels. In order to remove these non-ground features in the 3D voxels, a ground layer depth map for each image was needed to prune projected voxels beyond a threshold. In other words, we computed only associations for voxels close to the geometry of the ground layer depth map. It is worth pointing out that the ground layer depth map is not generated from the photogrammetry process (e.g., ContextCapture) but rendered using the segmented ground layer mesh and the camera parameters.

**Depth Pooling Layer**

As Figure 1 illustrated, multiple 2D feature maps are back-projected into the same voxel grids, and a feature selection mechanism is needed to select the appropriate feature for each 3D voxel. A pooling layer can be used since it is commonly used in a CNN architecture to reduce the spatial resolution of the hidden representations and aggregating information. The pooling layer in a 2D network has a pooling window with a fixed shape (e.g., *2*2*), and it is used to slide over all regions in the input according to its stride and to compute a single output for each location it traversed. Maximum pooling and average pooling are the two commonly used pooling operators. A max-pooling operator will take the maximum value of the region it operates on in the original input and create a new output matrix with the reduced size. An average pooling works the same way, but instead of using the maximum value, it will compute the region's average.

For 3DMV architecture, the purpose of using a pooling layer is not to reduce the spatial resolution of the 2D feature map but to extract the meaningful feature for projection. Thus, the pooling window has a shape of $1*1*n$, where $n$ is the number of the images that can be projected to the same voxel. The original 3DMV used a max-pooling layer that computed the maximum value on each feature channel over all associated 2D feature maps. The max-pooling strategy worked well when handling RGB-D sensed indoor data since the physical distance between each source image and the 3D objects was relatively short (e.g., 5 to 10 meters). In that case, the max-pooling layer simply carried forward only the largest information available amplitude-wise and did not consider the distance between the source image and the objects. However, for outdoor photogrammetric data, the physical distance between the source image and the 3D objects varied widely from frame to frame.

To collect aerial photos using UAV for covering a large area of interest, a UAV needs to fly at an altitude window of 70 to 400 meters. Depending on the time constraints of the data collection process and the resolution of the 3D reconstruction, the overlap between adjacent frames varies from 70% to 85%. Consequentially, the distance between different images to the same objects would have a large discrepancy. Furthermore, the distance should be considered at the pixel level rather than the image level since aerial photos can capture different objects with various distances. In addition, the same object can be captured at different angles, and better details when captured at an angle closer to 90 degrees. Thus, the angle between the images to the objects is an important consideration.

As shown in Figure 1, we designed a depth-pooling layer to replace the max-pooling layer in the original 3DMV architecture with the considerations mentioned above. The depth pooling layer is essentially a mask that is computed for each 3D voxel grid using all the depth images with which it is associated. If its depth value is the smallest among all pixels associated with that voxel from other images, the features in each pixel of the 2D feature map will be back-projected into the 3D voxel. The actual implementation of this depth-pooling layer will still use a max-pooling API call, but instead of operating on the feature map, we operate on the depth map with all values multiplied by -1. Note that selecting features from an image with an angle closer to 90 degrees to the objects was not embedded in the depth-pooling layer but was pre-computed when establishing the correspondence between the source images and the 3D voxels. The photo with a better angle (closer to 90 degrees) has a higher priority to be selected to associate with the voxel, and only three images are used to extract the 2D features for each voxel.

## GROUND MATERIAL DATABASE

In order to evaluate our modified 3DMV architecture with the depth-pooling layer, a large ground material database is needed. Note that 3D data needs to be reconstructed using aerial photos and photogrammetric techniques. To feed the data to the 3DMV model, both the 3D point clouds and the source images need to be annotated to create the database. A thorough search of the relevant literature does not yield any existing material database that meets the requirements. In our previous work that was published in 2019 (Chen et al., 2019), a relatively small orthophoto ground material database was created, which contained four data sets from the OWT repository. In this work, we projected the class labels from the orthophoto to the 3D point cloud and further expanded the database to include 16 more datasets. Our research team has put a large amount of effort into creating the database. We manually annotated the ground layer point clouds from our previously created point cloud database, which was annotated to have point-wise labels for man-made structures, vegetation, and ground (Chen et al., 2020a). Detailed information on the ground material database is summarized in Table 1.

**Table 1. UAV-based photogrammetric ground material database**

| *Name* | *Number of points* | *Bare earth* | *Road* | *Grass* | *Area size (sqkm)* | *Number of images* | *State* | *Data source* |
|---|---|---|---|---|---|---|---|---|
| AP Hill | 419,288 | 63% | 37% | 0% | 0.055 | 918 | VA | Marines |
| Camp Moreno | 1,346,394 | 89% | 10% | 1% | 0.179 | 1,876 | CA | Navy/Marines |
| Camp Roberts CACTF | 2,619,216 | 96% | 4% | 0% | 0.337 | 3,494 | CA | Navy/Marines |
| Catalina | 801,470 | 71% | 12% | 14% | 0.095 | 1,954 | CA | Civilian |
| IIT | 723,915 | 32% | 11% | 58% | 0.098 | 1,282 | CA | Marines |
| CTFMOUT | 704,839 | 35% | 5% | 60% | 0.090 | 1,268 | NC | Army |
| Ft Drum | 18,038,253 | 2% | 38% | 61% | 2.446 | 6,657 | NY | Army |
| Ft Pickett AssultCourse | 625,497 | 72% | 14% | 13% | 0.083 | 521 | VA | Army |
| Ft Pickett CACTF | 390,527 | 75% | 19% | 6% | 0.050 | 1,353 | VA | Army |
| Ft Pickett Classroom | 496,333 | 63% | 31% | 6% | 0.066 | 354 | VA | Army |
| Ft Pickett Mout Facility | 450,008 | 81% | 15% | 4% | 0.058 | 1,247 | VA | Army |
| Ft Pickett Range 21 | 3,195,411 | 47% | 4% | 49% | 0.418 | 707 | VA | Army |
| Lejeune MOUT | 2,976,618 | 63% | 17% | 19% | 0.418 | 3,104 | NC | Marines |
| McMillan | 757,415 | 56% | 30% | 14% | 0.098 | 1,541 | CA | Navy/Marines |
| MUTC | 5,883,478 | 10% | 15% | 74% | 0.962 | 2,843 | IN | Army |
| OCCC | 2,003,887 | 8% | 53% | 38% | 0.271 | 1,824 | FL | Civilian |
| Range 230 | 672,653 | 91% | 9% | 0% | 0.091 | 1,451 | CA | Marines |
| USAFA | 902,193 | 78% | 9% | 13% | 0.121 | 1,230 | CA | Air Force |
| USC | 2,082,856 | 1% | 59% | 40% | 0.419 | 10,588 | CA | Civilian |
| Apartments | 306,411 | 1% | 84% | 15% | 0.043 | 1,970 | CA | Civilian |

A total of 20 datasets were used to create our ground material database which covers approximately 6.4 square kilometers of the area with about 45 million points. These datasets were collected from different geographic locations within the U.S. for various operational military purposes. Due to the data acquisition time constraints, the total number of collected images varies across different datasets ranging from a few hundred to ten thousand. Since this study focused on segmenting point clouds that cover a large area of interest, a high point density required a longer model training and testing time. Thus, all point clouds were down-sampled with 0.3-meter point spacing. The point clouds were manually labeled with the following three labels: (1) bare earth, (2) roads, and (3) grass. The road category consisting of paved road points, sidewalk points, and any man-made or artificial surface points. Note that the entire database composed of 52% bare earth, 24% road, and 24% grass. Using this database as the training set, the trained model was biased towards the bare earth. Thus, we used the histogram of classes represented in the database to weigh the loss during training to overcome this data imbalance issue. Figure 2 shows an example of the point cloud and its annotations in our created database.

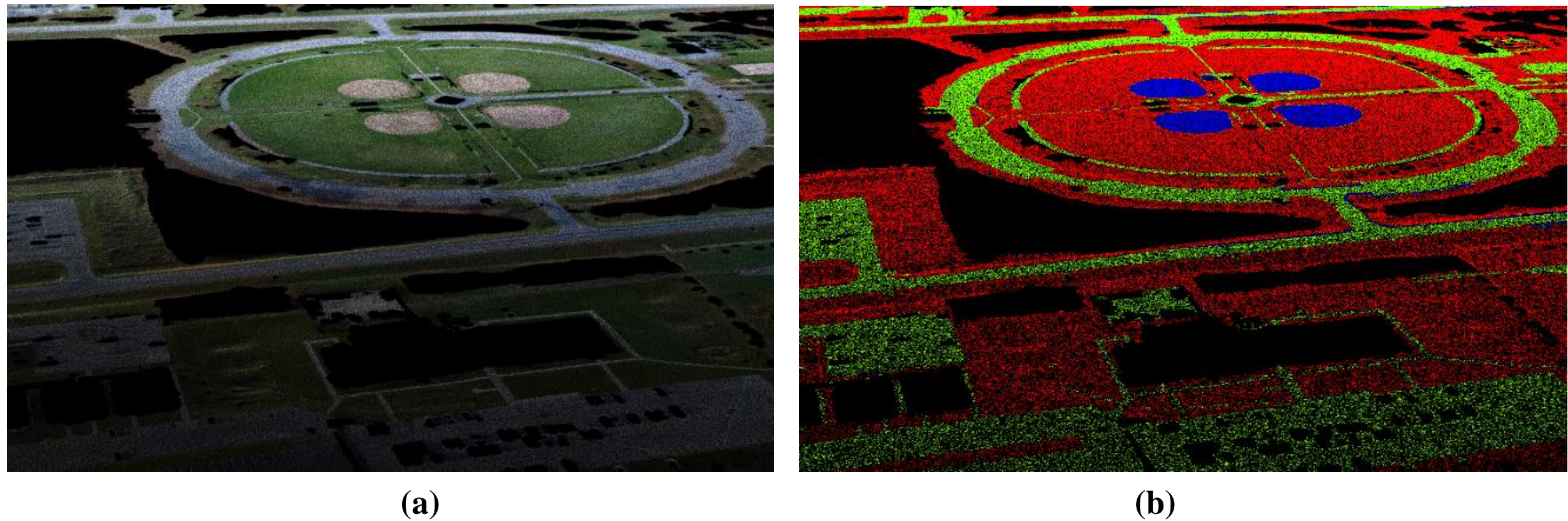

**(a)** **(b)**

**Figure 2. Example data set in the database. (a) Colorized ground point cloud. (b) The annotated point cloud, where the bare earth, road, and grass are marked with blue, green, and red, respectively.**

As previously mentioned, to train the 3DMV model, not only the 3D data need to be annotated, the 2D source images also need to be labeled. As can be seen from Table 1, there are a total of 46,182 images have been collected and used for the 3D reconstruction process. Manually annotating all the source images will be an extremely time-consuming and labor-intensive task. Fortunately, with the existing correspondence between the source images and the 3D data, the point cloud labels can be back-projected onto the images and create the ground-truth label mask. Figure 3 shows a few examples of the source images and their associated label masks. Note that buildings and trees are not shown in the label mask since they have already been removed from the 3D point cloud and are not needed for this work.

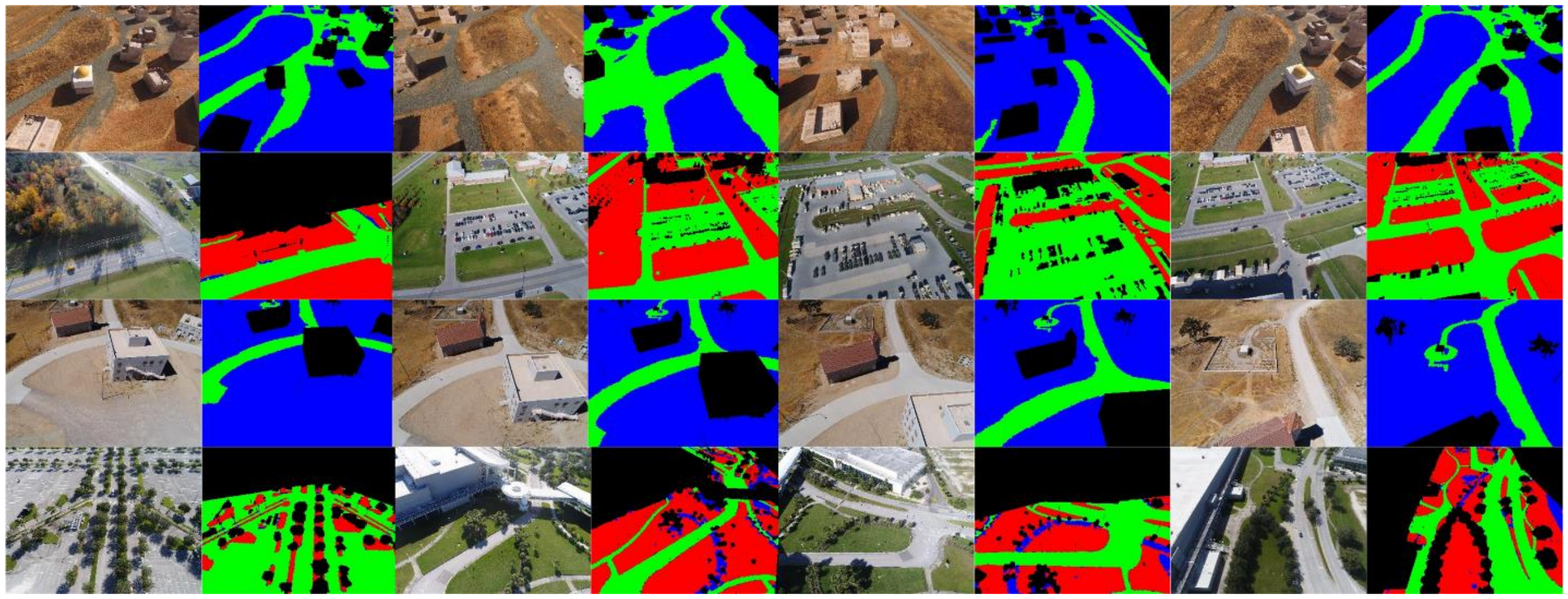

**Figure 3. Examples of the annotated source images.**

## EXPERIMENT AND RESULTS

In this experiment, all models were trained using the datasets discussed in the previous section and tested on a new dataset that the models had never seen before. The main objectives of conducting the experiment in this study were 1) comparing the ground material segmentation performance using only the 2D source images and using both 2D and 3D information, and 2) evaluating our designed depth-pooling layer compared to the original max-pooling layer in the 3DMV model. Three models were trained in the experiment. For the first model, we only trained the 2D network and directly projected the 2D segmentation results onto the 3D point cloud. For the second and third models, we trained the original 3DMV model with the max-pooling layer and our designed depth-pooling layer, respectively. The commonly used precision, recall, harmonic mean of precision and recall (f1 score), and intersection over union (IOU)—also known as the Jaccard Index—were all used to evaluate the segmentation results. Segmentation results are summarized in Tables 2 to 4.

**Table 2. Experimental results for using 2D network only.**

| | precision | recall | f1-score | IOU. |
|---|---|---|---|---|
| bare earth | 0.93 | 0.24 | 0.38 | 0.23 |
| road | 0.32 | 0.93 | 0.48 | 0.31 |
| grass | 0.45 | 0.79 | 0.58 | 0.40 |
| macro avg | 0.57 | 0.65 | 0.48 | 0.32 |
| weighted avg | 0.73 | 0.47 | 0.44 | 0.28 |

Using the 2D network for the segmentation task, the model was able to achieve only a weighted average f1-score of 0.44 and IOU of 0.28. When we used the original 3DMV model, the segmentation results improved by 0.16 on the f1 score and 0.15 on the IOU. Finally, by using our designed depth-pooling layer in the 3DMV network, we were able to achieve the highest f1 score of 0.78 and IOU of 0.64. From table 2 we can see that when we only used the 2D network, bare earth was unable to be retrieved, as indicated by the low recall value of 0.24. Since it only relied on the color information, the model was confused between the soil and yellow grass.

**Table 3. Experimental results for using 2D+3D networks.**

| | precision | recall | f1-score | IOU. |
|---|---|---|---|---|
| bare earth | 0.86 | 0.47 | 0.61 | 0.44 |
| road | 0.36 | 0.90 | 0.51 | 0.34 |
| grass | 0.59 | 0.71 | 0.64 | 0.47 |
| macro avg | 0.60 | 0.69 | 0.59 | 0.42 |
| weighted avg | 0.72 | 0.59 | 0.60 | 0.43 |

**Table 4. Experimental results for using 2D+3D networks with depth pooling layer.**

| points | precision | recall | f1-score | IOU. |
|---|---|---|---|---|
| bare earth | 0.92 | 0.72 | 0.81 | 0.68 |
| road | 0.59 | 0.91 | 0.72 | 0.56 |
| grass | 0.69 | 0.84 | 0.76 | 0.61 |
| macro avg | 0.73 | 0.83 | 0.76 | 0.62 |
| weighted avg | 0.82 | 0.78 | 0.78 | 0.64 |

The result was improved when we added the 3D information during the segmentation process by using the original 3DMV. The recall was almost doubled (0.47), and the precision of the grass was increased by 0.14. The result was further improved when we took the distance between the source images and the ground into consideration and used the depth-pooling layer. The precision was largely improved from 0.36 to 0.59 while the recall of the road was almost unchanged (0.91 in table 4 compared to 0.9 in table 3). This is because roads usually have a narrow shape, meaning better 2D features can be extracted using the source images when captured closer to the roads. Overall, we can clearly see the benefits of adding 3D information for the task and taking into consideration the distance between the source images and the ground. Figure 4 shows the result of the labeled ground point cloud using our modified 3DMV model.

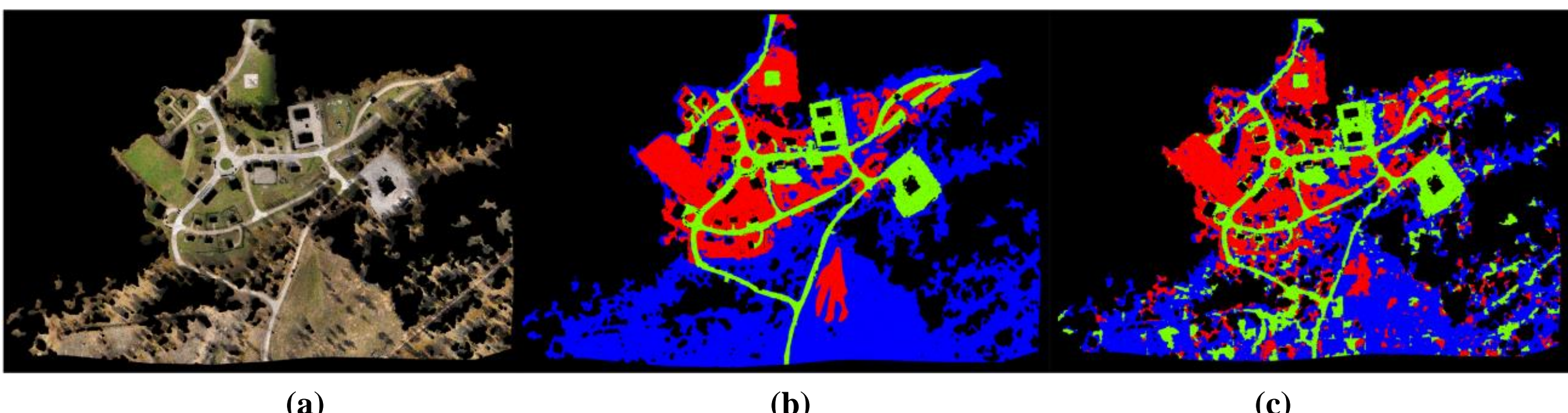

**(a)** **(b)** **(c)**

**Figure 4. Ground material segmentation result using 3DMV with depth pooling layer. (a) Ground layer point cloud, (b) ground truth, and (c) prediction result.**

## Creating Virtual Environment Using the Segmented Photogrammetric Data and the Extracted Ground Materials

We utilized the Unreal4 game engine (UE4) to create a simulation environment with the segmented photogrammetric data and the extracted ground material information for proof of concept. To make the virtual environments, we used our previously designed STPLS+ to segment the photogrammetric meshes into top-level terrain elements (i.e., ground, man-made structures, and vegetation). For more information on STPLS+, the readers can refer to (Chen et al., 2019; Chen, McAlinden, Spicer, & Soibelman, 2019; Chen, Feng, McAlinden, & Soibelman, 2020; Chen et al., 2020a; Chen et al., 2020b). Following that, ground materials were segmented using the work presented in this paper.

The following data needed to be prepared for creating the virtual environment in the game engine. (1) 3D mesh for the man-made structures in the OBJ format, (2) a digital elevation model (DEM) in GeoTIFF format, (3) a texture map for the ground, (4) a mask image for the forests, (5) a mask image for the grass, and (6) a mask image for the bare earth. The DEM was easily created using the segmented ground layer data with open-sourced software such as CloudCompare (Girardeau-Montaut, 2016). The ground layer orthophoto was rendered using the segmented 3D point cloud and was used as the ground texture map. The forest mask was generated by projecting the segmented 3D tree data onto the x-y plane. Similarly, the grass and bare earth masks were generated by projecting the segmented ground material point clouds onto the x-y plane.

Man-made structure meshes were easily imported into UE4 as static mesh actors. To import all other above-mentioned data into UE4, we used a third-party plugin TerraForm and the Procedural Landscape Ecosystem, which is commercially available. TerraForm was designed to import geographical data into UE4 and place useful assets and actors. The primary reason for using TerraForm in this study was that the DEM GeoTIFF data could be imported into UE4 through TerraForm as a landscape actor. The main benefit of creating a landscape actor in UE4 instead of creating a static mesh actor for the ground is that landscape actor provides a higher performance and makes features such as painting, laying trees, and placing grass easier. Procedural Landscape Ecosystem is a package obtained from the UE4 marketplace for placing trees, grass, and bare earth on landscape actors with the created mask images. For the sake of demonstration, a video demo of driving inside the Camp Grayling area was created and can be found at https://www.youtube.com/watch?v=SKhMhF-ZyxM. Figure 5 shows two examples of the Camp Grayling virtual environment demo.

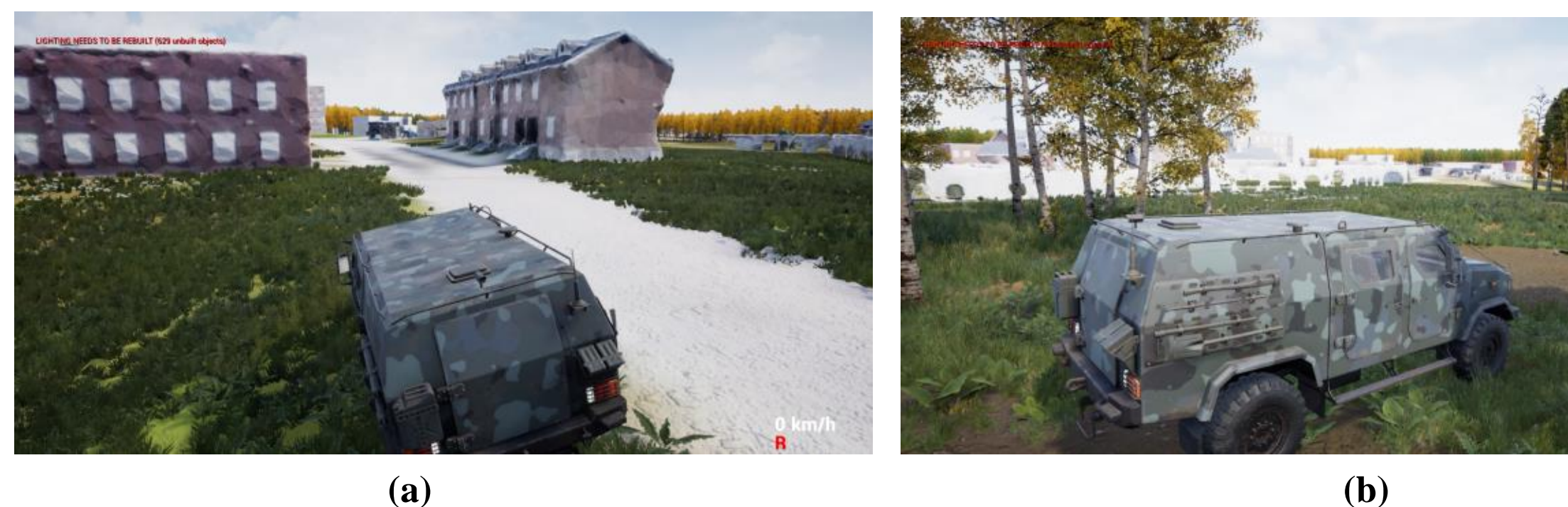

**(a)** **(b)**

**Figure 5. The created Camp Grayling virtual environment**

## CONCLUSION

It is undeniable that the aerial photogrammetry technique revolutionized how photorealistic 3D models are being produced for creating virtual simulation environments. One of the main challenges of using the raw photogrammetric data for simulation has been that it does not contain any semantic information/features of objects. This study has expanded our previous work and further investigated the ground material segmentation problem using a 2D-3D hybrid approach. An improved 3DMV deep-learning architecture with our designed depth-pooling layer was proposed, and a large aerial photogrammetric ground material database containing 20 data sets was created. An experiment was conducted to evaluate our improved 3DMV model. The results showed that the model that used both 2D and 3D information for segmenting ground materials could outperform the model that only used the 2D data (i.e., the f1 score was improved by 0.16, and the IOU was improved by 0.15). In addition, replacing the original max-pooling layer with our designed depth-pooling layer in the 3DMV architecture further enhanced the result and achieved an f1 score of 0.78 and IOU of 0.64. With the identified ground materials and the segmented photogrammetric data, we also recommend two packages to create a virtual environment in UE4. Finally, a video demonstration of the virtual Camp Grayling area with the added ground covers was provided for proof of concept.

## ACKNOWLEDGMENTS

The authors would like to thank the two primary sponsors of this research: Army Futures Command (AFC), Synthetic Training Environment (STE), and the Office of Naval Research (ONR). They would also like to acknowledge the assistance provided by 3/7 Special Forces Group (SFG), Naval Special Warfare (NSW), the National Training Center (NTC), and the U.S. Marine Corps (USMC). This work is supported by University Affiliated Research Center (UARC) award W911NF-14-D-0005. Statements and opinions expressed and content included do not necessarily reflect the position or the policy of the Government, and no official endorsement should be inferred.

## REFERENCES

Adelson, E. H. (2001). On seeing stuff: The perception of materials by humans and machines. Paper presented at the *Human Vision and Electronic Imaging VI, , 4299* 1-13.

Chen, M., Feng, A., McAlinden, R., & Soibelman, L. (2020). Photogrammetric point cloud segmentation and object information extraction for creating virtual environments and simulations. *Journal of Management in Engineering, 36*(2), 04019046.

Chen, M., Feng, A., McCullough, K., Prasad, P. B., McAlinden, R., & Soibelman, L. (2020a). 3D photogrammetry point cloud segmentation using a model ensembling framework. *Journal of Computing in Civil Engineering, 34*(6), 04020048.

Chen, M., Feng, A., McCullough, K., Prasad, P. B., McAlinden, R., & Soibelman, L. (2020b). Generating synthetic photogrammetric data for training deep learning based 3D point cloud segmentation models. *arXiv Preprint arXiv:2008.09647,*

Chen, M., McAlinden, R., Spicer, R., & Soibelman, L. (2019). Semantic modeling of outdoor scenes for the creation of virtual environments and simulations. Paper presented at the *Proceedings of the 52nd Hawaii International Conference on System Sciences,*

Chen, M., Feng, A., McCullough, K., Prasad, P. B., McAlinden, R., Soibelman, L., & Enloe, M. (2019). Fully automated photogrammetric data segmentation and object information extraction approach for creating simulation terrain. Paper presented at the *Interservice/Industry Training, Simulation, and Education Conference (I/ITSEC),*

Cimpoi, M., Maji, S., & Vedaldi, A. (2015). Deep filter banks for texture recognition and segmentation. Paper presented at the *2015 IEEE Conference on Computer Vision and Pattern Recognition (CVPR),* 3828-3836.

Dai, A., & Nießner, M. (2018). 3dmv: Joint 3d-multi-view prediction for 3d semantic scene segmentation. Paper presented at the *Proceedings of the European Conference on Computer Vision (ECCV),* 452-468.

Dana, K. J., Van Ginneken, B., Nayar, S. K., & Koenderink, J. J. (1999). Reflectance and texture of real-world surfaces. *ACM Transactions on Graphics (TOG), 18*(1), 1-34.

Girardeau-Montaut, D. (2016). CloudCompare. *Retrieved from CloudCompare: Https://Www.Danielgm.Net/Cc,*

Liu, C., Sharan, L., Adelson, E. H., & Rosenholtz, R. (2010). Exploring features in a bayesian framework for material recognition. Paper presented at the *Computer Vision and Pattern Recognition (CVPR), 2010 IEEE Conference On,* 239-246.

Liu, L., & Fieguth, P. (2012). Texture classification from random features. *IEEE Transactions on Pattern Analysis and Machine Intelligence, 34*(3), 574-586.

Paszke, A., Chaurasia, A., Kim, S., & Culurciello, E. (2016). Enet: A deep neural network architecture for real-time semantic segmentation. *arXiv Preprint arXiv:1606.02147,*

Sharan, L., Rosenholtz, R., & Adelson, E. (2009). Material perception: What can you see in a brief glance? *Journal of Vision, 9*(8), 784-784.

Varma, M., & Zisserman, A. (2009). A statistical approach to material classification using image patch exemplars. *IEEE Transactions on Pattern Analysis and Machine Intelligence, 31*(11), 2032-2047.

Zeiler, M. D., & Fergus, R. (2014). Visualizing and understanding convolutional networks. Paper presented at the *European Conference on Computer Vision,* 818-833.